%% file: arxiv.tex
\documentclass{article}

\usepackage[sort,authoryear,round]{natbib}
\usepackage{fullpage}

\usepackage[utf8]{inputenc} 
\usepackage[T1]{fontenc}    
\usepackage{xcolor}
\usepackage{hyperref}       
\definecolor{myblue}{rgb}{0,0.2,0.8}
\hypersetup{ %
pdftitle={},
pdfauthor={},
pdfsubject={},
pdfkeywords={},
pdfborder=0 0 0,
pdfpagemode=UseNone,
colorlinks=true,
linkcolor=myblue,
citecolor=myblue,
filecolor=myblue,
urlcolor=myblue,
pdfview=FitH}
\usepackage{authblk}
\usepackage{url}            
\usepackage{booktabs}       
\usepackage{amsfonts}       
\usepackage{amsthm}
\usepackage{nicefrac}       
\usepackage{microtype}      
\usepackage{amsmath} 

\usepackage{graphicx}
\usepackage{xspace}
\usepackage[varqu,varl,var0,scaled=0.97]{inconsolata}
\usepackage{caption}
\usepackage{overpic}
\usepackage{wrapfig}
\usepackage{tcolorbox}
\usepackage{enumitem}
\usepackage{threeparttable}

\usepackage{tikz}
\usepackage{subfigure}
\usepackage{xspace}
\usepackage{hyperref}
\usepackage{listings}
\usepackage{array}
\usepackage{url}
\usepackage{breakurl}

\usepackage{listings}
\usepackage{color}
\usepackage{lipsum}

\definecolor{dkgreen}{rgb}{0,0.6,0}
\definecolor{gray}{rgb}{0.5,0.5,0.5}
\definecolor{mauve}{rgb}{0.58,0,0.82}

\newcommand\cnns{CNNs\xspace}
\newcommand\batchnorm{BN\xspace}

\newcommand\dnns{DNNs\xspace}
\newcommand\flops{FLOPs\xspace}
\newcommand\fullfinetuning{FFT\xspace}
\newcommand\singularvaluedecomp{SVD\xspace}
\newcommand\peft{PEFT\xspace}

\newcommand\vit{ViT\xspace}
\newcommand\visualwakewords{VWW\xspace}
\newcommand\llms{LLMs\xspace}
\newcommand\lora{\textsc{LoRA}\xspace}
\newcommand\dora{\textsc{DoRA}\xspace}
\newcommand\galore{\textsc{GaLore}\xspace}
\newcommand\dsc{DSC\xspace}

\newcommand\dscs{DSCs\xspace}
\newcommand\bnh{BN+H\xspace}
\newcommand\tta{TTA\xspace}

\newcommand{\eg}{\emph{e.g.},\xspace}
\newcommand{\ie}{\emph{i.e.},\xspace}

\usepackage{subcaption}

\newcommand{\figref}[1]{Fig.~\ref{#1}}

\title{\bf{From LLMs to Edge: \\ Parameter-Efficient Fine-Tuning on Edge Devices}}
\author[1]{Georg Slamanig$^*$}
\author[1]{Francesco Corti$^*$}
\author[1,2]{Olga Saukh}
\affil[1]{Graz University of Technology, Austria}
\affil[2]{CSH Vienna, Austria}
\affil[ ]{\texttt{\{georg.slamanig@student. francesco.corti@,saukh@\}tugraz.at}}

\begin{document}
\date{}
\maketitle

\def\thefootnote{*}\footnotetext{Equal contribution}
\renewcommand{\thefootnote}{\P}
\begin{abstract}
Parameter-efficient fine-tuning (\peft) methods reduce the computational costs of updating deep learning models by minimizing the number of additional parameters used to adapt a model to a downstream task. While extensively researched in large language models (\llms), their application to smaller models used on edge devices, such as convolutional neural networks, remains underexplored. This paper benchmarks and analyzes popular \peft methods on convolutional architectures typically deployed in resource-constrained edge environments. We evaluate \lora, \dora, and \galore for updating standard and depthwise convolutional architectures to handle distribution shifts and accommodate unseen classes. We utilize recently proposed PyTorch profilers to compare the updated model performance and computational costs of these \peft methods with traditional fine-tuning approaches. With resource efficiency in mind, we investigate their update behavior across different rank dimensions. We find that the evaluated \peft methods are only half as memory-efficient when applied to depthwise-separable convolution architectures, compared to their efficiency with \llms.
Conversely, when targeting convolutional architectures optimized for edge deployment, adapter-based \peft methods can reduce floating point operations (\flops) during model updates by up to 95\%. These insights offer valuable guidance for selecting \peft methods based on hardware constraints, performance requirements, and application needs. Our code is online\footnote{\url{https://github.com/gslama12/pytorch-model-profiler}}.
\end{abstract}
\renewcommand{\thefootnote}{\arabic{footnote}}

\section{Introduction}
  \label{sec:intro}

Edge devices are increasingly used to deploy deep neural network (DNN) models for local data processing~\citep{zhou2019edge}, enhancing application accessibility and security by reducing latency and preserving user privacy through the elimination of server communication~\citep{shi2016edge}. However, these devices face significant hardware constraints, such as limited memory, computational capacity and dynamic resource constraints~\citep{corti2023reds}, which pose challenges for deploying \dnns~\citep{lin2023tiny}. Additionally, \dnns deployed at the edge often struggle with distribution shifts in incoming data, leading to degraded performance over time~\citep{quinonero2008dataset}. Maintaining performance under such conditions requires models to be updated efficiently within the hardware constraints specific to each edge device. 

Several approaches have been proposed to update models for unseen classes and distribution shifts. These fall into two categories: methods that build robust invariant models, and those that use domain adaptation techniques. The former pre-train models to be resilient to distribution shifts using data augmentations~\citep{shorten2019survey}, contrastive loss functions~\citep{khosla2020supervised}, and regularization strategies that reduce sensitivity to domain discrepancies~\citep{arjovsky2019invariant}. The latter adapt models at deployment time by training a parameterized subspace of configurable networks~\citep{saukh2023representing} or fine-tuning pre-trained features via backpropagation~\citep{rumelhart1986learning}. While invariant models are more robust to expected shifts, they may fail to generalize to unseen distributions and can be brittle when adapted via backpropagation~\citep{arjovsky2019invariant}.

However, on-device adaptation is limited by the computational constraints of edge devices, since standard backpropagation requires at least three times more computation than inference~\citep{xu2022mandheling}. This challenge is exacerbated by the high resource demands of state-of-the-art models. For instance, Vision Transformers (\vit), despite their strong performance in computer vision~\citep{dosovitskiy2021an}, have quadratic computational complexity with respect to input size, making them unsuitable for resource-constrained settings~\citep{dosovitskiy2021an}. As a result, Convolutional Neural Networks (\cnns), whose complexity scales linearly with input size~\citep{howard2017mobilenets}, are typically preferred for edge deployment. Still, updating \cnns via backpropagation remains computationally expensive and often impractical on low-power devices~\citep{han2016eie}. To address this, MobileNet architectures replace standard convolutions with depthwise-separable convolutions (\dscs), reducing inference computation by a factor of 8 to 9~\citep{howard2017mobilenets}.

\begin{figure*}[htbp]
  \centering
  \subfigure[MobileNetV2]{
    \includegraphics[width=0.24\textwidth, trim={0.6cm 0cm 5.5cm 0cm},clip]{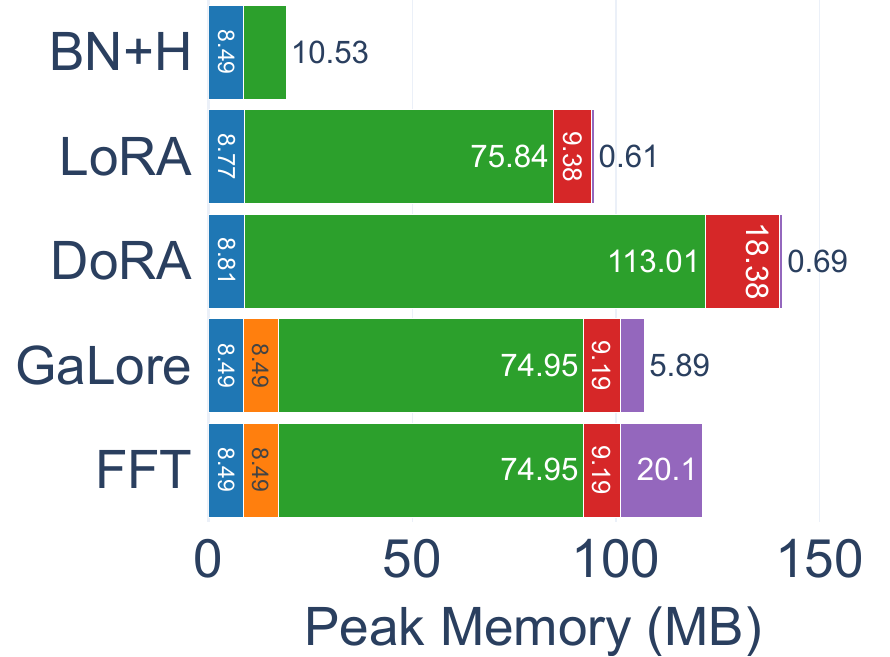}
    \label{fig:mem_mnv2}
  }
  \subfigure[MobileNetV3]{
    \includegraphics[width=0.24\textwidth, trim={0.6cm 0cm 5.5cm 0cm},clip]{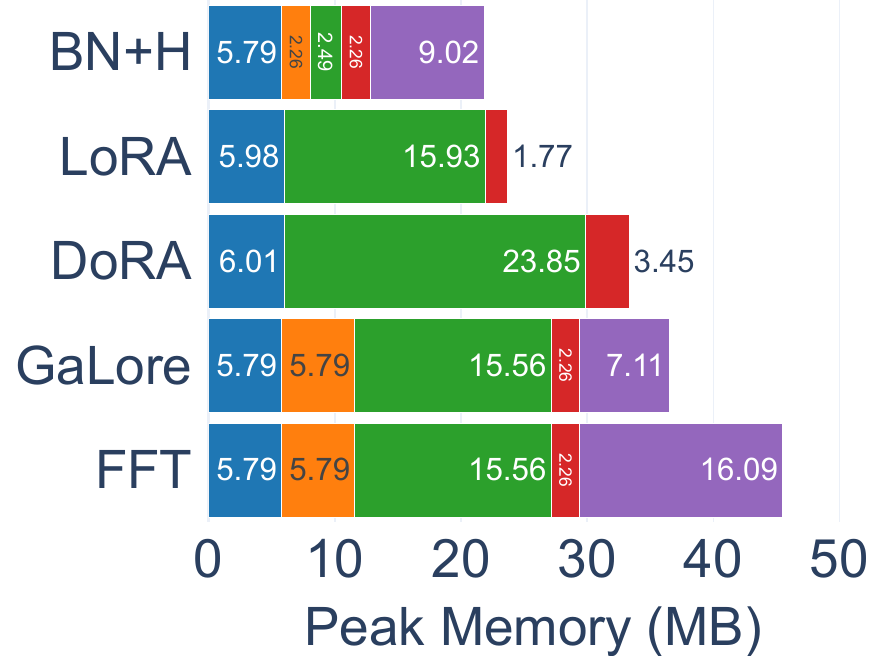}
    \label{fig:mem_mnv3}
  }
  \subfigure[ResNet-18]{
    \includegraphics[width=0.33\textwidth, trim={0.6cm 0cm 0cm 0cm},clip]{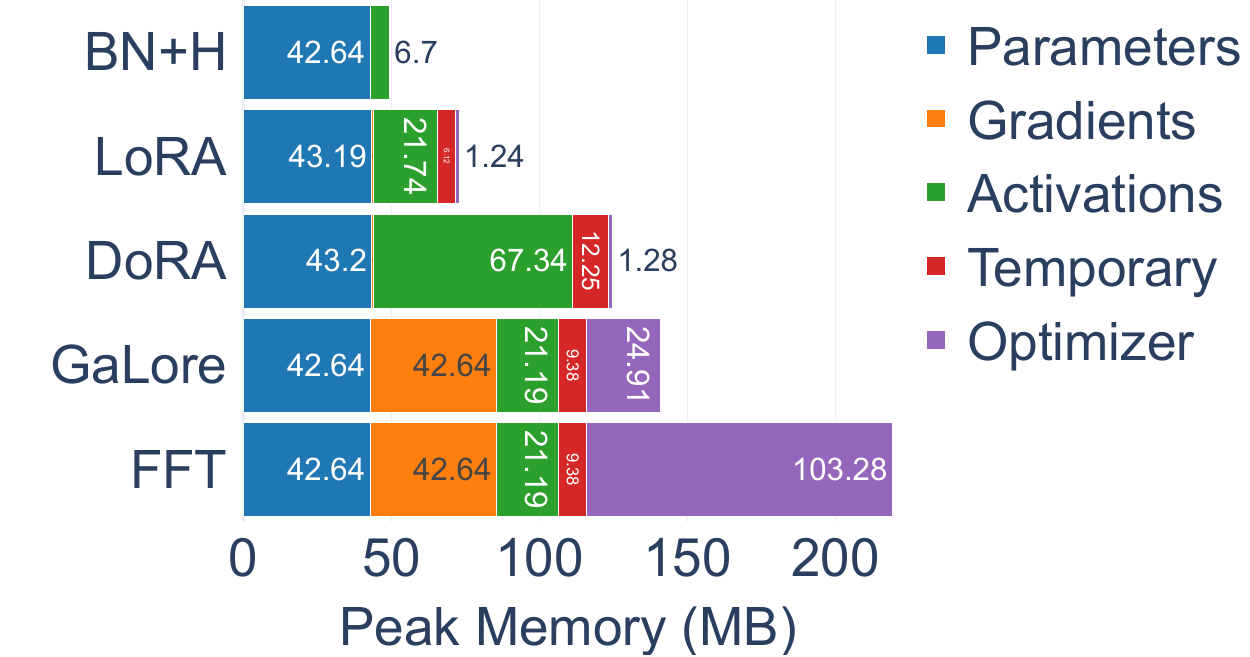}
    \label{fig:mem_rn18}
  }
  \caption{\textbf{Peak memory consumption analysis of \peft methods for different models.} Analysis of forward and backward passes for a single $224\times224$ image. The model architecture influences the total peak memory across the profiled memory groups. For depthwise convolution models, \lora, \dora, \galore, and \bnh show higher peak memory usage compared to standard convolution models due to activations required memory. The peak memory usage of \dora is 49\% to 51\% higher than that of \lora for the MobileNetV2 depthwise architecture. For ResNet18 a larger share of gradient calculation and optimizer state memory enhances the efficiency of \lora and \dora \peft methods.}
  \label{fig:mem}
\end{figure*}

To overcome the limitations of updating models on edge devices with constrained resources, parameter-efficient fine-tuning (\peft) methods have emerged as a promising solution. Assuming updates to a pre-trained model lie in a low-rank subspace~\citep{hu2021lora}, \peft reduces computational and memory demands by restricting gradients or weight updates to low-rank representations~\citep{hu2021lora, zhao2024galorememoryefficientllmtraining}. While effective in LLMs~\citep{han2024parameter}, \peft remains underexplored for \cnns in edge vision tasks. Moreover, tools for evaluating \peft on \cnns in terms of resource usage, complexity, and accuracy are lacking.

This paper addresses the question: \emph{Given the hardware constraints of an edge device and a pre-trained convolutional model, which \peft method best enables efficient and effective on-device updates?} We propose a novel framework to estimate the efficiency and effectiveness of \peft methods for specific tasks. Our contributions are: 
\begin{itemize}
    \item We extend PyTorch’s FLOPs counter and memory tracker to assess \peft methods on CNN models typically deployed on edge devices.
    \item We evaluate \peft performance on pre-trained \cnns, analyzing peak memory usage, \flops, and accuracy when updating to unseen classes or handling distribution shifts.
    \item We benchmark \lora, \dora, and \galore for updating standard and depthwise convolutional architectures to handle distribution shifts and accommodate unseen classes. We show that fine-tuning depthwise convolutional architectures can diminish resource efficiency of \peft methods.
    \item We investigate the impact of the rank hyperparameter on the adaptation accuracy.
\end{itemize}

\section{Parameter-Efficient Fine-Tuning}
\label{sec:peft}
\peft methods reduce computational and memory costs by minimizing the number of updated parameters. This is achieved by assuming that weight updates lie in a low-rank subspace, leading to lightweight adaptations without sacrificing performance~\citep{hu2021lora}. While extensively studied in \llms, the application of \peft methods to \cnns under resource constraints remains underexplored. This section introduces popular \peft methods evaluated in this work: Low-Rank Adaptation (\lora), Weight-Decomposed Low-Rank Adaptation (\dora), Gradient Low-Rank Projection (\galore), and head-only fine-tuning with batch-normalization (\bnh).

\textbf{Low-Rank Adaptation (\lora).}
\lora introduces low-rank matrices into the weight update process, enabling fine-tuning with a small number of additional parameters~\citep{hu2021lora}. Specifically, the weight update $\Delta W$ is expressed as $\Delta W = A B^T$, where $A \in \mathbb{R}^{d \times r}$ and $B \in \mathbb{R}^{d \times r}$ are low-rank matrices with $r \ll d$. This decomposition significantly reduces the number of parameters and computational complexity, as the gradients are computed and applied only for the low-rank factors $A$ and $B$. Advantages of \lora include its simplicity, scalability, and compatibility with various architectures. However, its performance may degrade when the rank $r$ is set too low, particularly for tasks requiring significant adaptation.

\begin{figure*}[htbp]
  \centering
  \subfigure[MobileNetV2]{
    \includegraphics[width=0.24\textwidth, trim={0.9cm 0cm 5cm 0cm},clip]{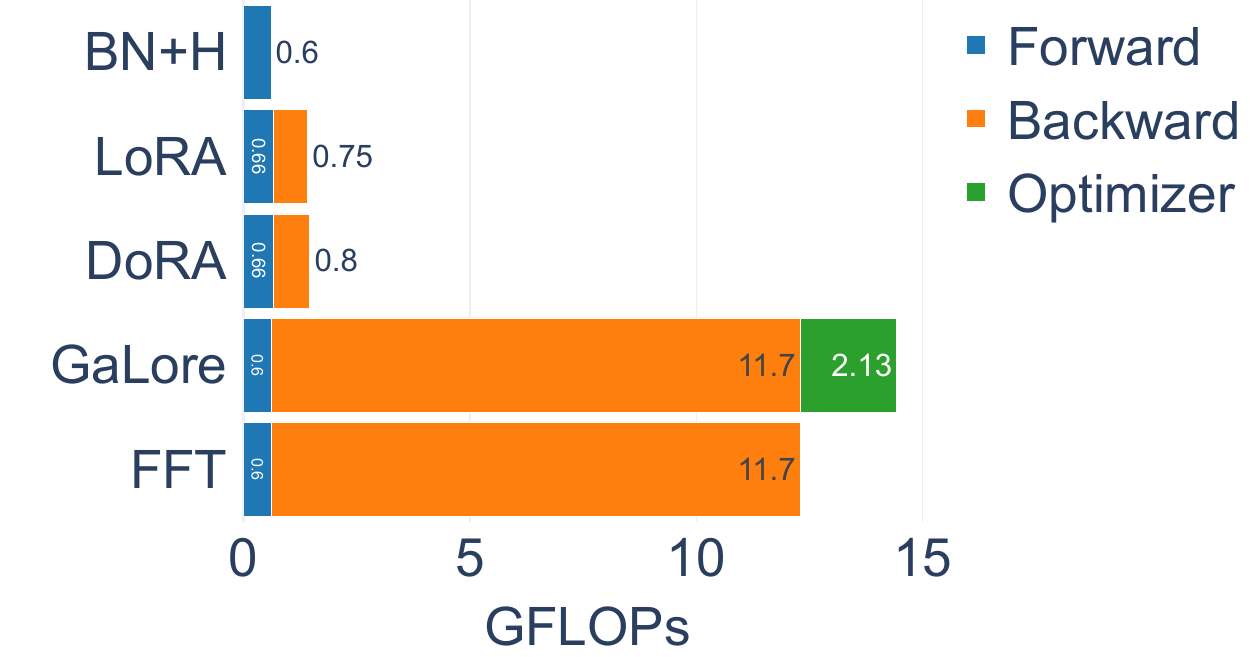}
    \label{fig:flops_mnv2}
  }
  \subfigure[MobileNetV3]{
    \includegraphics[width=0.24\textwidth, trim={0.9cm 0cm 5.2cm 0cm},clip]{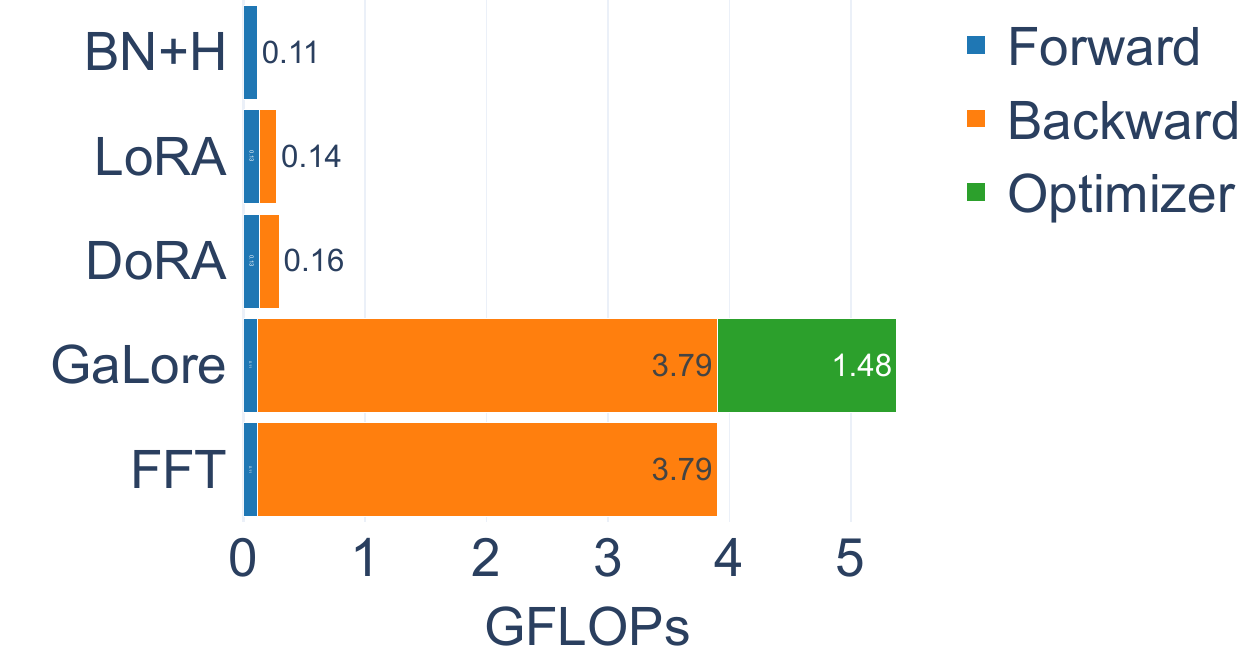}
    \label{fig:flops_mnv3}
  }
  \subfigure[ResNet-18]{
    \includegraphics[width=0.33\textwidth, trim={0.9cm 0cm 0cm 0cm},clip]{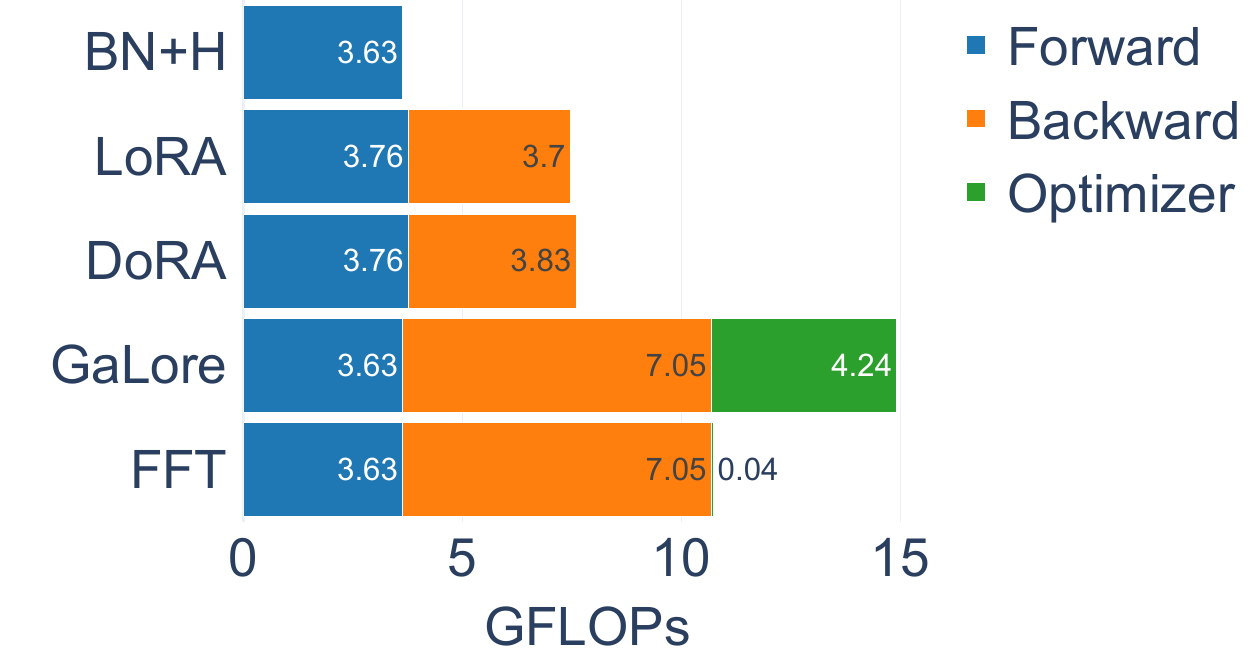}
    \label{fig:flops_rn18}
  }
  \caption{\textbf{\flops analysis of \peft methods for different models.} Analysis of forward and backward passes for a single $224\times224$ image. Except for \galore all the \peft methods reduce the required \flops significantly compared to \fullfinetuning. Depthwise architectures (\ie MobileNets) report a \flops reduction of more than  10$\times$  compared to standard convolution.}
   \label{fig:flops}
\end{figure*}

\textbf{Weight-Decomposed Low-Rank Adaptation (\dora).}
\dora initially decomposes the pre-trained weights $W_0$ into magnitude vector $m$ and direction matrix $V$ and only applies \lora to $V$. During training, only $m$ and $V$ are updated. The fine-tuned weights $W'$ can be formulated as $W' = m \frac{V + \Delta V}{V + \Delta V} = m \frac{W_0 + BA}{||W_0 + BA||_c}$ where $\Delta V = BA$ is the directional update and $||\cdot||_c$ represents the vector-wise norm across each column vector~\cite{liu2024dora}. The authors introduce \dora as an alternative to \lora, demonstrating that it more closely resembles the training behavior of full fine-tuning when comparing the magnitude and directional updates of the weight matrices of \llms during training~\citep{liu2024dora}. The trainable magnitude vectors introduce slightly more trainable parameters for \dora compared to \lora. Additionally, the weight decomposition in \dora introduces a more complex computational graph during backpropagation.

\textbf{Gradient Low-Rank Projection (\galore).}
\galore combines low-rank approximation with gradient sensitivity to adaptively refine parameter updates~\citep{zhao2024galorememoryefficientllmtraining}. Given the gradient matrix $G$, \galore formulates updates as $G \approx \sum_{i=1}^r \sigma_i u_i v_i^T$ where $u_i$ and $v_i$ are singular vectors and $\sigma_i$ are singular values obtained from Singular Value Decomposition (\singularvaluedecomp) of $G$. \galore dynamically determines the rank $r$ by truncating based on a threshold $\epsilon$, ensuring $\sum_{i=1}^r \sigma_i / \sum_{j} \sigma_j \geq 1 - \epsilon$. This approach ensures an optimal trade-off between computational efficiency and approximation accuracy, adapting resource usage to the task requirements. However, the reliance on \singularvaluedecomp and rank truncation can be computationally expensive, especially for high-dimensional gradients.

\textbf{Head-only Fine-Tuning with Batch-Normalization (\bnh).}
This fine-tuning approach only updates the final classification layer (\ie head) using backpropagation and the batch normalization (\batchnorm) statistics during the forward pass. The updates of the \batchnorm statistics introduce additional flexibility and help to address feature distribution shifts, improving update performance in certain scenarios~\citep{frankle2020training}. While the low number of trainable parameters makes this method highly resource efficient, its effectiveness depends on the degree of distribution shift between the pre-training and target datasets. 

We evaluate the practicality of \peft methods for \cnns on edge devices by analyzing trade-offs in computation, memory, and accuracy, and compare them to head-only with batch normalization updates (\bnh) and full fine-tuning (\fullfinetuning).  

\section{Profiling \peft Methods}
\label{sec:profiling}
\textbf{Performance Measures.} We evaluate the performance of \peft methods by measuring the number of \flops and the peak memory usage required to update the models to each task during the forward and backward passes. The former serves as the inference latency to estimate the model's execution time on an edge device~\citep{liberis2021munas}, while the latter is considered the major bottleneck for enabling neural networks on the edge~\citep{lin2023tiny}. Using these measurements for each \peft method configuration, along with the edge device's hardware and latency constraints, our framework offers guidance to estimate the efficiency and effectiveness of \peft methods for specific tasks.

\subsection{Profiling Framework}
We modify an existing \flops counter\footnote{\url{https://gist.github.com/soumith/5f81c3d40d41bb9d08041431c656b233}} to distinguish between the \flops used during gradient computation and those used during the optimizer's weight update step in the backward pass. An example usage can be seen in Listing~\ref{code:flops_counter}.

\begin{lstlisting}[%
  caption=\textbf{Example of profiler usage to compute model's forward-backward \flops.}, 
  label=code:flops_counter,
  xleftmargin=1em,
  xrightmargin=1em]
flops_counter = FlopCounterMode(model)

with flops_counter:
    optimizer.zero_grad()
    outputs = model(input_tensor)
    loss = outputs.sum()
    loss.backward()
    flops_counter.reset_module_tracking()
    optimizer.step()
\end{lstlisting}

The \flops counter uses \texttt{\_\_torch\_dispatch\_\_} to attach hooks to the tensor level operations of PyTorch~\citep{paszke2019pytorch}. By tracking the operations for tensor convolution, multiplication, addition and batch normalization we calculate the number of required \flops for each operation from the operand shapes.

\begin{figure*}[htbp]
  \subfigure[MobileNetV2]{
    \includegraphics[width=0.265\textwidth, trim={0cm 0cm 5cm 0cm},clip]{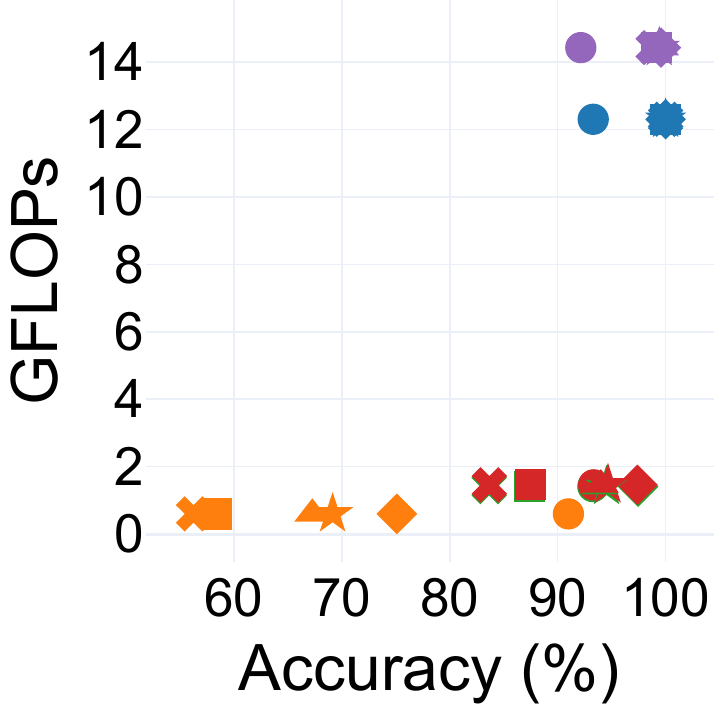}
    \label{fig:acc-vs-res-a}
  }
  \subfigure[MobileNetV3]{
    \includegraphics[width=0.265\textwidth, trim={0cm 0cm 5cm 0cm},clip]{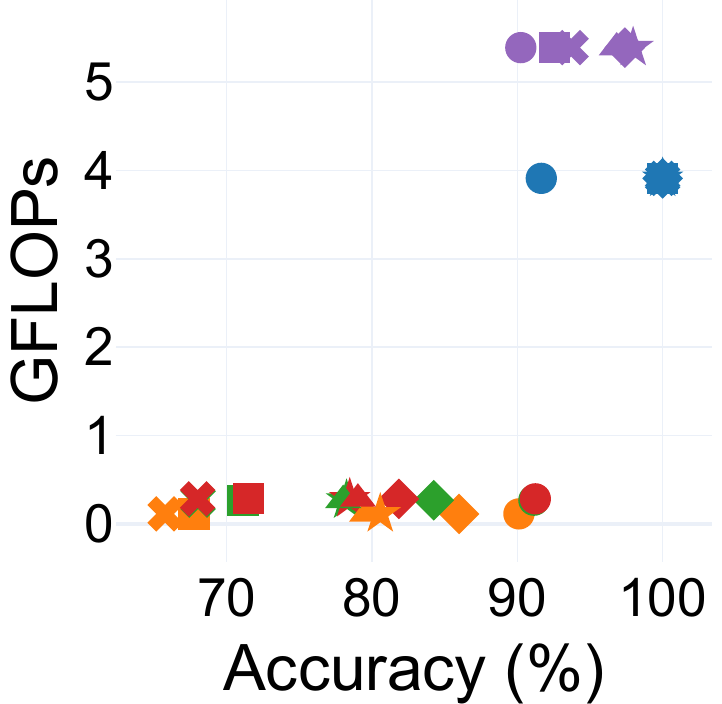}
    \label{fig:acc-vs-res-c}
  }
  \subfigure[ResNet-18]{
    \includegraphics[width=0.265\textwidth, trim={0cm 0cm 5cm 0cm},clip]{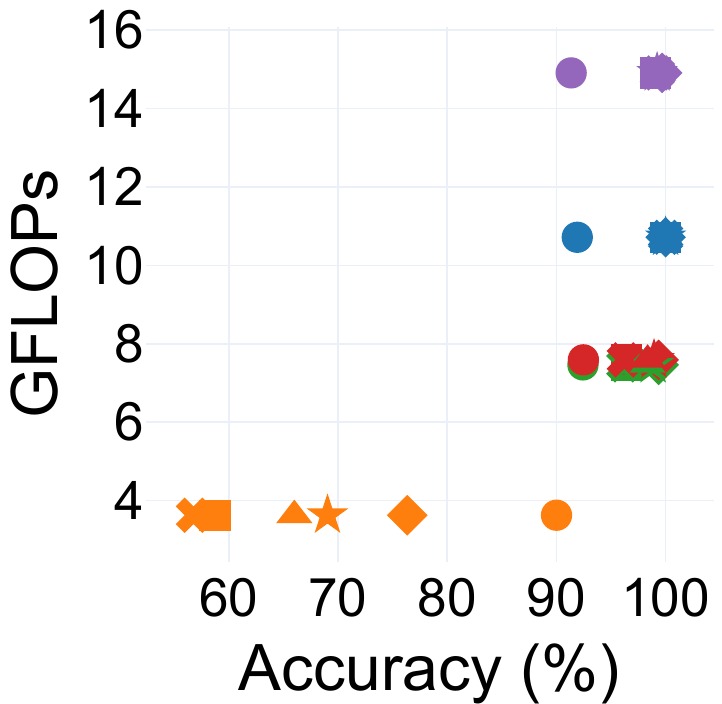}
    \label{fig:acc-vs-res-d}
  }
  \subfigure{
        \includegraphics[height=3.0cm, trim={12.4cm 1cm 0cm 0cm},clip]{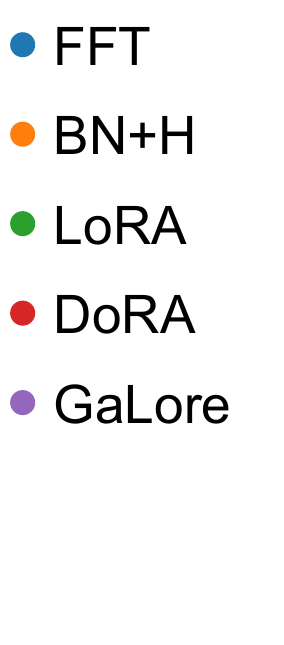}
  }
  \\
  \subfigure[MobileNetV2]{
    \includegraphics[width=0.265\textwidth, trim={0cm 0cm 5cm 0cm},clip]{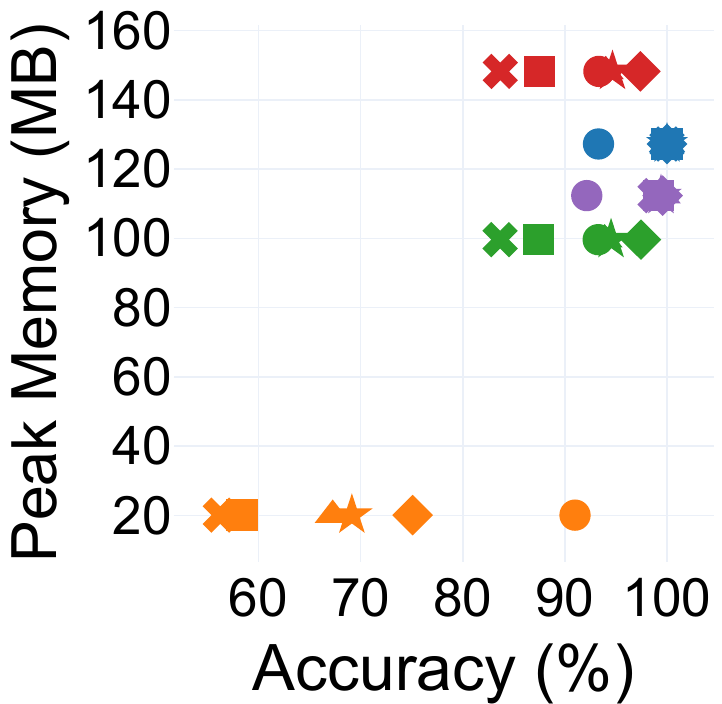}
    \label{fig:acc-vs-res-e}
  }
  \subfigure[MobileNetV3]{
    \includegraphics[width=0.265\textwidth, trim={0cm 0cm 5cm 0cm},clip]{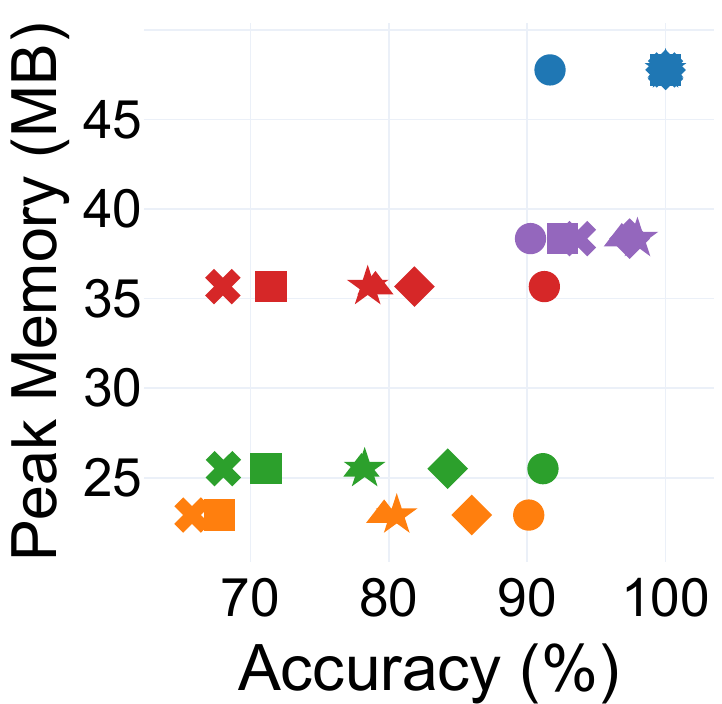}
    \label{fig:acc-vs-res-g}
  }
  \subfigure[ResNet-18]{
    \includegraphics[width=0.265\textwidth, trim={0cm 0cm 5cm 0cm},clip]{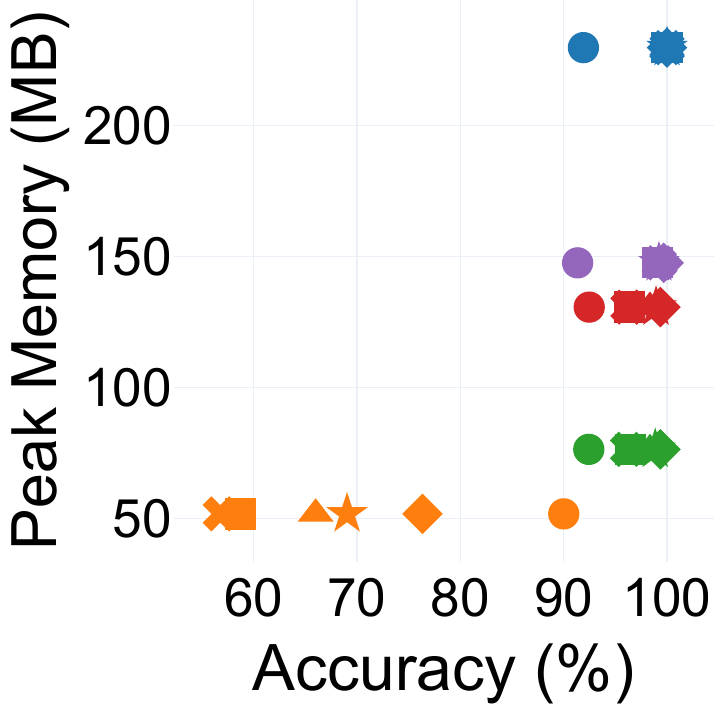}
    \label{fig:acc-vs-res-h}
   }
   \subfigure{
    \label{fig:acc-vs-res-i}
        \includegraphics[height=3.0cm, trim={11.3cm 1cm 0cm 0cm},clip]{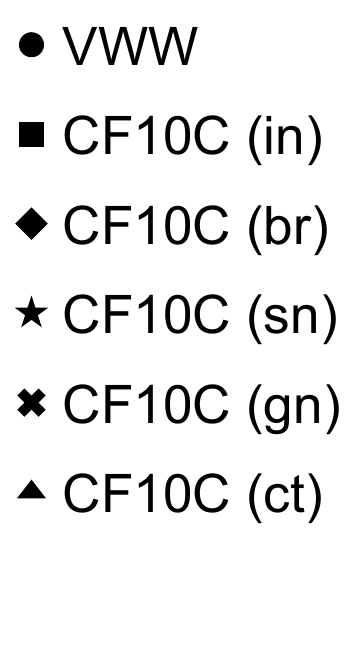}
    }
    \caption{\textbf{Trade-off between accuracy and resource usage.} Profiling and evaluation of the \peft methods on four different models pre-trained on ImageNet, across different fine-tuning tasks for one training step using a single $224\times224$ image. While \galore achieves consistent accuracy results comparable to \fullfinetuning across different models and fine-tuning tasks, \lora and \dora show accuracy variations of up to 20\% across tasks on MobileNet architectures. \lora delivers the best trade-off between accuracy and resource consumption across various tasks on ResNet-18. Accuracies are reported using early stopping after 10 epochs without validation loss improvement.}
    \label{fig:acc-vs-res}
\end{figure*}

\textbf{Memory.} PyTorch's latest memory tracker\footnote{\url{https://github.com/pytorch/pytorch/pull/124688}} can distinguish between the memory groups listed in Table \ref{tab:mem-context}. We modify the memory tracker to profile the peak memory usage of each group regardless of when it occurs, with total memory computed as their sum and profiling done using the steps reported in Listing~\ref{code:flops_counter}.

\begin{table}[htbp]
  \centering
  \caption{\textbf{PyTorch memory groups analyzed during training.}
   The tracker categorizes memory usage by group, offering detailed insights into resource allocation.}
  \begin{tabular}{{lp{0.80\linewidth}}}
    \toprule
    \textbf{Group} & \textbf{Description} \\
    \midrule
    PARAM & Model parameters. \\ 
    GRAD  & Gradients of model weights during backpropagation. \\ 
    ACT   & Intermediate activations stored for the backward pass. \\ 
    OPT   & Optimizer state memory (\eg momentum buffers). \\ 
    TEMP  & Temporary buffers used in gradient computations (\eg autograd intermediates). \\
    \bottomrule
  \end{tabular}
  \label{tab:mem-context}
\end{table}

\section{Evaluation}
\label{sec:evaluation}
To investigate the capabilities of \peft methods for models typically deployed on edge devices, we evaluate the performance of \lora, \dora, \galore, \bnh and \fullfinetuning on MobileNetV2~\citep{sandler2018mobilenetv2} and MobileNetV3~\citep{howard2019searching}. These models employ optimized \dscs layers, which reduce the computational cost during inference by up to $9\times$ times~\citep{howard2017mobilenets}. To highlight the performance and efficiency differences of \peft methods on \dsc architectures compared to standard convolution architectures, we also include the results for ResNet-18~\citep{he2016deep}.
We show accuracy and profiling results for all models, initially pre-trained on ImageNet~\citep{deng2009imagenet}, on various downstream tasks, including CIFAR10-C~\citep{hendrycks2019benchmarking} corruptions and the Visual Wake Words~\citep{chowdhery2019visual} dataset (\visualwakewords). We choose CIFAR-10-C corruptions of varying difficulty to highlight the task-specific achieved accuracy of the investigated \peft methods and demonstrate the advantages of \lora, \dora, and \galore over the simpler \bnh approach. Furthermore, we evaluate the effectiveness of the selected \peft methods on the \visualwakewords dataset for the binary classification task of detecting the presence of a person in an image. Additionally, we conduct the same experiments on the MobileNetV2 and ResNet18 models pre-trained on CIFAR-10~\citep{cifar100}.
For all the experiments we followed the hyperparameter recommendations from~\citep{hu2021lora, liu2024dora, zhao2024galorememoryefficientllmtraining} to ensure a fair comparison, see Table~\ref{tab:hyperparams}. We use implementations of \lora and \dora from the Hugging Face \peft library\footnote{\url{https://huggingface.co/docs/peft/index}} and the pre-release implementation of \galore\footnote{\url{https://github.com/jiaweizzhao/GaLore}} according to~\citet{zhao2024galorememoryefficientllmtraining} with small adaptions to suit CNN architectures on the edge. 

\textbf{Memory.} In \figref{fig:mem} we analyze the peak memory usage of the models' forward pass, backward pass, and optimizer step for all \peft methods. We observe that the investigated \peft methods only reduce memory usage for gradients and optimizer groups. For models that use \dscs, the storage required for activation maps is the primary contributor to total peak memory, limiting the effectiveness of \peft methods for such model architectures.
With minimal trainable parameters, \bnh fine-tuning stands out as the most memory-efficient \peft method. By avoiding full backpropagation through all layers, it achieves up to 85\% total peak memory savings compared to \fullfinetuning (\figref{fig:mem_mnv2}). Since the classifier of MobileNetV3 comprises two trainable linear layers, \bnh exhibits reduced memory efficiency on this model compared to MobileNetV2, achieving only a 52\% reduction in peak memory usage relative to \fullfinetuning (\figref{fig:mem_mnv3}).

We observe \lora to be the second most memory efficient \peft method, with a peak memory reduction of up to 67\% compared to \fullfinetuning on ResNet-18 (\figref{fig:mem_rn18}), replicating the improvement observed on \llms~\citep{hu2021lora}.  However, for models using \dscs, we observe a smaller peak memory reduction, ranging between 22\% on MobileNetV2 (\figref{fig:mem_mnv2}) and 48\% on MobileNetV3 (\figref{fig:mem_mnv3}), which is notably lower than the improvements observed in \llms. Standard convolution \cnns, such as ResNet-18, allocate a larger portion of peak memory to optimizer state and gradients compared to models with \dscs, making \lora more memory-efficient for these architectures (\figref{fig:mem}).
Although \dora offers an almost costless alternative to \lora during inference~\citep{liu2024dora}, its more complex computational graph leads to a memory overhead between 29\% on MobileNetV3 (\figref{fig:mem_mnv3}) and 58\% on ResNet-18 (\figref{fig:mem_rn18}) during training.

Contrary to \lora and \dora, \galore only optimizes the memory used for storing optimizer states~\citep{zhao2024galorememoryefficientllmtraining}. This also implies that standard convolution \cnns, like ResNet-18, benefit more from \galore than models using \dscs, due to the larger share of optimizer state memory in the total peak memory (\figref{fig:mem_rn18}). Similarly, as reported in~\citet{zhao2024galorememoryefficientllmtraining} for \llms, our results in \figref{fig:mem} demonstrate an average reduction of 65\% in optimizer state memory across the tested models. However, for models using \dscs, this only results in an overall peak memory reduction of around 5-10\%, as the optimizer state memory is not the primary contributor (\figref{fig:mem_mnv2} and \figref{fig:mem_mnv3}). While ~\citet{zhao2024galorememoryefficientllmtraining} reports that \galore is more memory-efficient than \lora for \llms, we did not observe this behavior in the \cnns evaluated in this study (\figref{fig:mem}). With its significantly smaller number of trainable parameters, \lora utilizes only about 10\% of the optimizer state memory required by \galore, resulting in higher memory efficiency. We observe similar results to those in \figref{fig:mem} for models pre-trained on CIFAR-10 using $32\times32$ images, remaining within acceptable limits.

\textbf{FLOPs.} In \figref{fig:flops} we analyzed the \flops required by the \peft methods to perform forward and backward passes. For standard convolution \cnns, the ratio of \flops required for the backward pass to those needed for the forward pass is approximately 2:1~\citep{epoch2021backwardforwardFLOPratio}. 
\dscs alter this ratio by splitting the filters of the convolutional layer into groups, where the number of groups equals the number of input channels $C_{\text{in}}$. With each filter only processing a single input channel, the \flops during the forward pass of \dsc layers are reduced by the factor of $\frac{1}{C_{\text{in}}}$. During the backward pass of \dsc layers, the \flops include computations for both the input gradient and the weight gradient. While the weight gradient benefits from reduced \flops due to filter grouping, the input gradient does not. Consequently, this results in an approximate 20:1 ratio between the \flops required for the forward pass and the backward pass during \fullfinetuning (\figref{fig:flops}). This finding underscores the need for further investigation into optimizing backward pass \flops for \dsc models. 

Adapter-based \peft methods, such as \lora and \dora, significantly reduce the 20:1 \flops ratio of \dscs to approximately 1.2:1, achieving an overall \flops reduction of 80\% for MobileNetV3 (\figref{fig:flops_mnv3}) compared to \fullfinetuning. For standard convolutional models like ResNet-18, which exhibit the expected 2:1 \flops ratio between the forward and backward pass, the \flops reduction achieved by \lora and \dora is limited to approximately 57\% (\figref{fig:flops_rn18}).
Additionally, we observe that the \singularvaluedecomp operation in \galore's optimizer step introduces a \flops overhead of 10\% to 30\% compared to \fullfinetuning (\figref{fig:flops}), slightly exceeding the 10\% overhead reported for \llms in~\citet{zhao2024galorememoryefficientllmtraining} in certain models.

\textbf{Accuracy and Performance.} Our results in \figref{fig:acc-vs-res} demonstrate, that the accuracy after fine-tuning with \peft methods is significantly influenced by the model architecture, the size of the model, and the specific fine-tuning task. 
While \bnh demonstrates acceptable performance on lightweight adaptions like the CIFAR10-C Brightness (br) corruption and the VWW dataset, \lora, \dora and \galore consistently outperform \bnh fine-tuning on all other tasks, with accuracy differences of up to 40\%. 

By utilizing full-rank weight updates, \galore shows the most consistent accuracy across different fine-tuning tasks and model architectures and achieves comparable results to \fullfinetuning. While \lora and \dora achieve similar accuracy scores on ResNet-18, the adapter-based \peft methods exhibit inconsistent performance across different fine-tuning tasks on MobileNet architectures. Notably, for challenging fine-tuning tasks such as Impulse noise (in) and Gaussian noise (gn), MobileNets pre-trained on ImageNet experience accuracy drops of up to 20\% when using \lora or \dora compared to \galore (\figref{fig:acc-vs-res-a} and \figref{fig:acc-vs-res-c}).
\galore offers greater robustness than \lora but uses 1.13–2$\times$ more memory and 2–20$\times$ more FLOPs, depending on the model. In contrast, \lora requires about 2$\times$ more training iterations to converge.

\figref{fig:acc-vs-res-g} shows that, unlike results for LLMs reported in~\citet{zhao2024galorememoryefficientllmtraining}, \lora outperforms \galore by up to 2.5\% on some tasks. Similarly, contrary to~\citet{liu2024dora}, \dora shows no accuracy gain over \lora on edge-optimized \cnns in any experiment (\figref{fig:acc-vs-res}). Results on CIFAR-10 pretraining mirror these trends, with all models improving on corruption tasks and \lora consistently offering the best accuracy-efficiency trade-off.

\begin{figure*}[htbp]
\centering
\subfigure[ImageNet $\rightarrow$ VWW]{
    \includegraphics[width=0.27\textwidth, trim={0cm 0cm 4.3cm 0cm},clip]{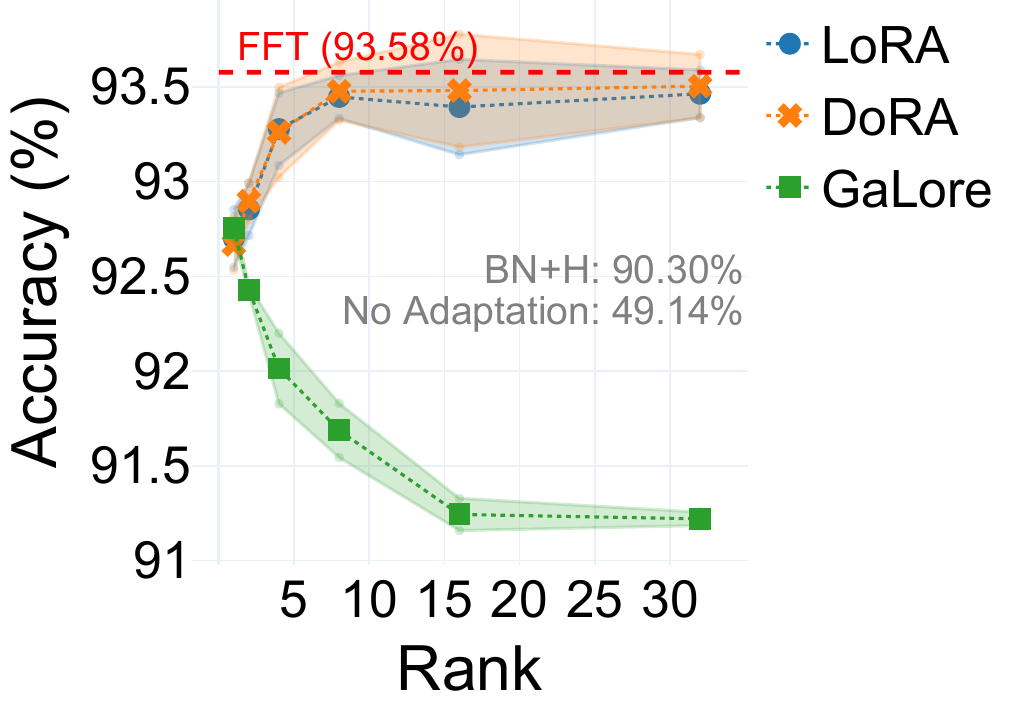}
    \label{fig:rank-acc-mnv2-1-a}
}
\subfigure[CIFAR-10 $\rightarrow$ VWW]{
    \includegraphics[width=0.27\textwidth, trim={0cm 0cm 4.3cm 0cm},clip]{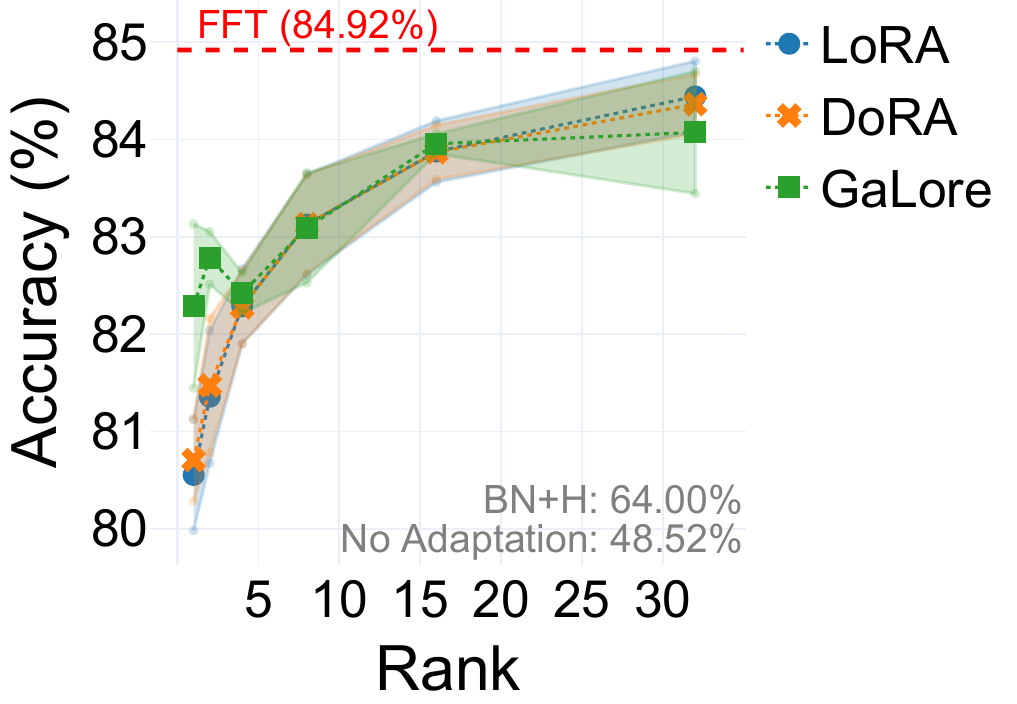}
    \label{fig:rank-acc-mnv2-1-b}
}
\subfigure[ImageNet $\rightarrow$ CIFAR-10-C (in)]{
    \includegraphics[width=0.36\textwidth, trim={0cm 0cm 0cm 0cm},clip]{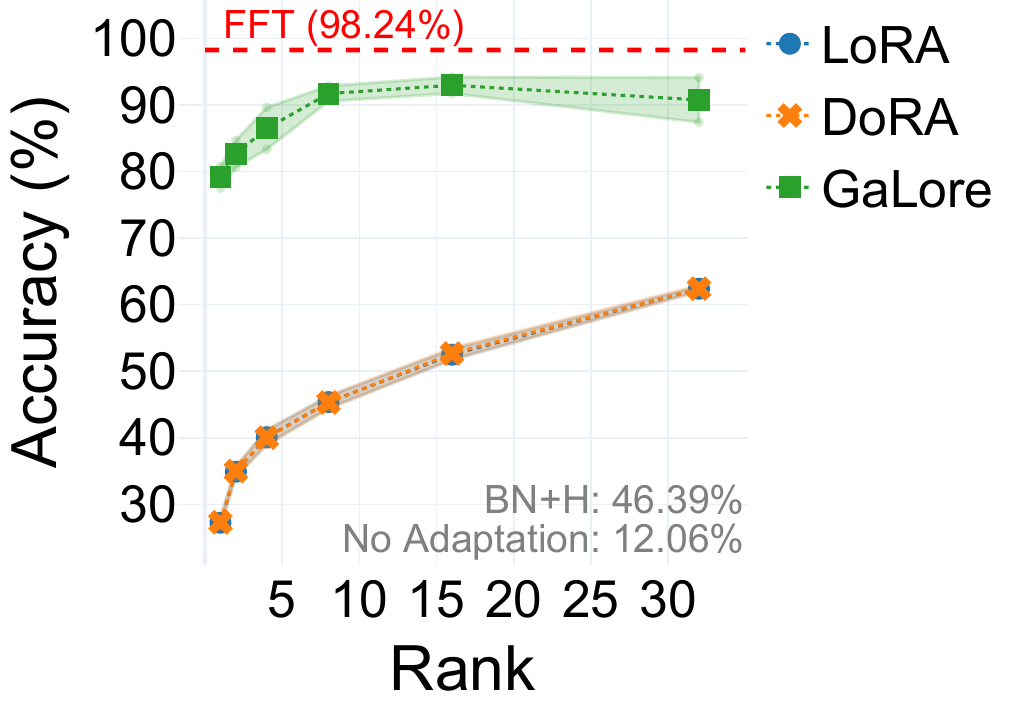}
    \label{fig:rank-acc-mnv2-1-c}
}
\\[\baselineskip]
\subfigure[CIFAR-10 $\rightarrow$ CIFAR-10-C (in)]{
    \includegraphics[width=0.27\textwidth, trim={0cm 0cm 4.3cm 0cm},clip]{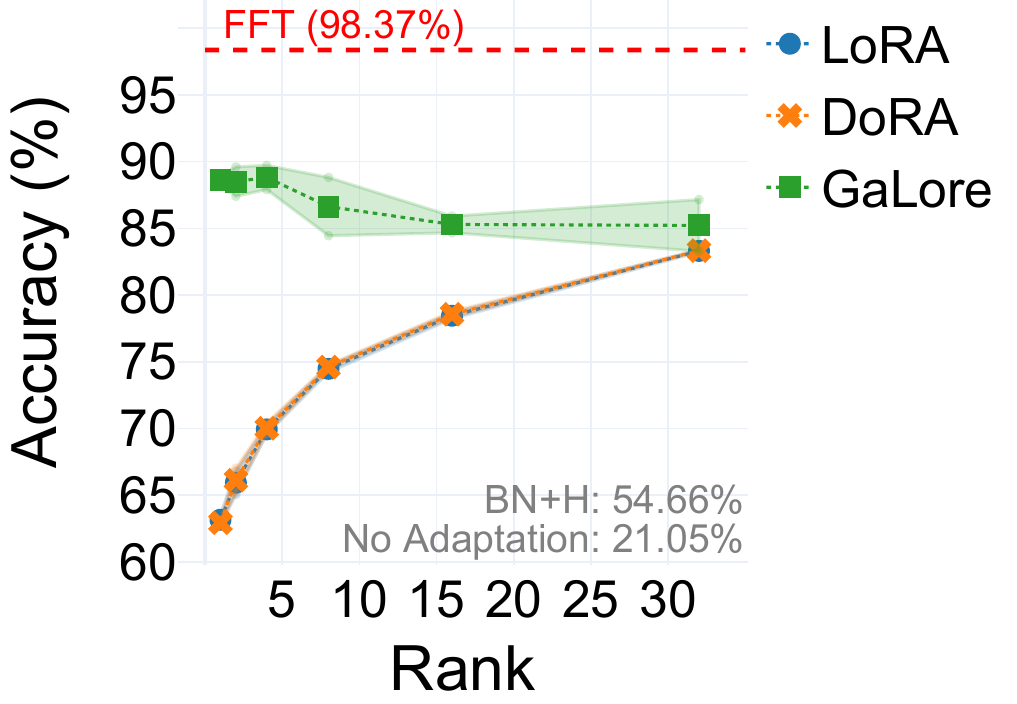}
    \label{fig:rank-acc-mnv2-1-d}
}
\subfigure[ImageNet $\rightarrow$ CIFAR-10-C (br)]{
    \includegraphics[width=0.27\textwidth, trim={0cm 0cm 4.3cm 0cm},clip]{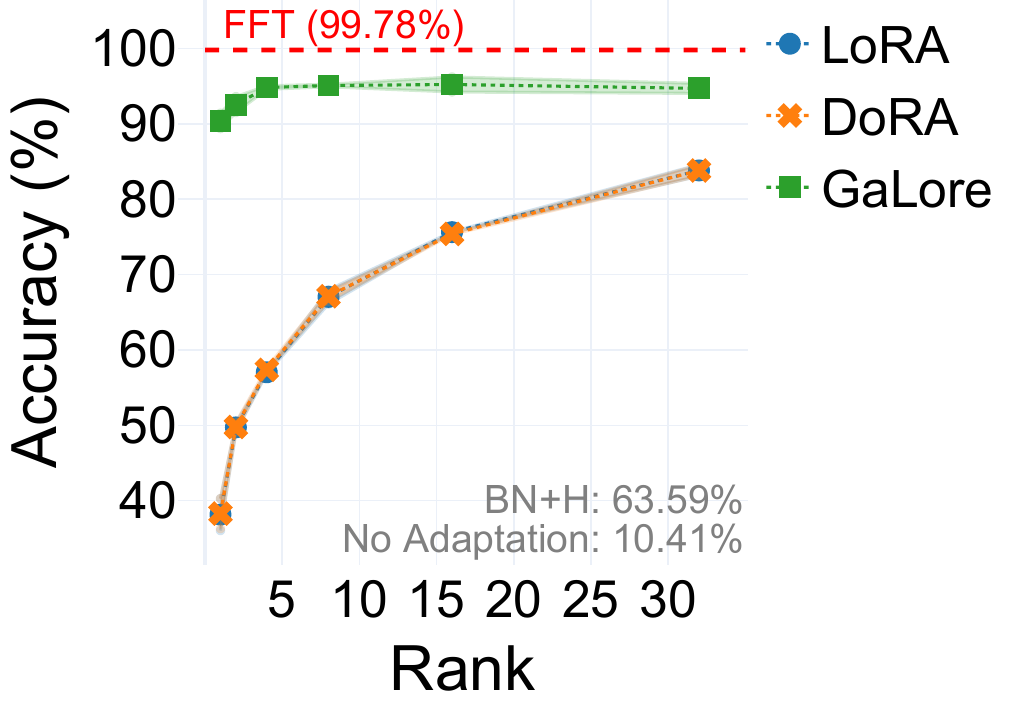}
    \label{fig:rank-acc-mnv2-1-e}
}
\subfigure[CIFAR-10 $\rightarrow$ CIFAR-10-C (br)]{
    \includegraphics[width=0.36\textwidth, trim={0cm 0cm 0cm 0cm},clip]{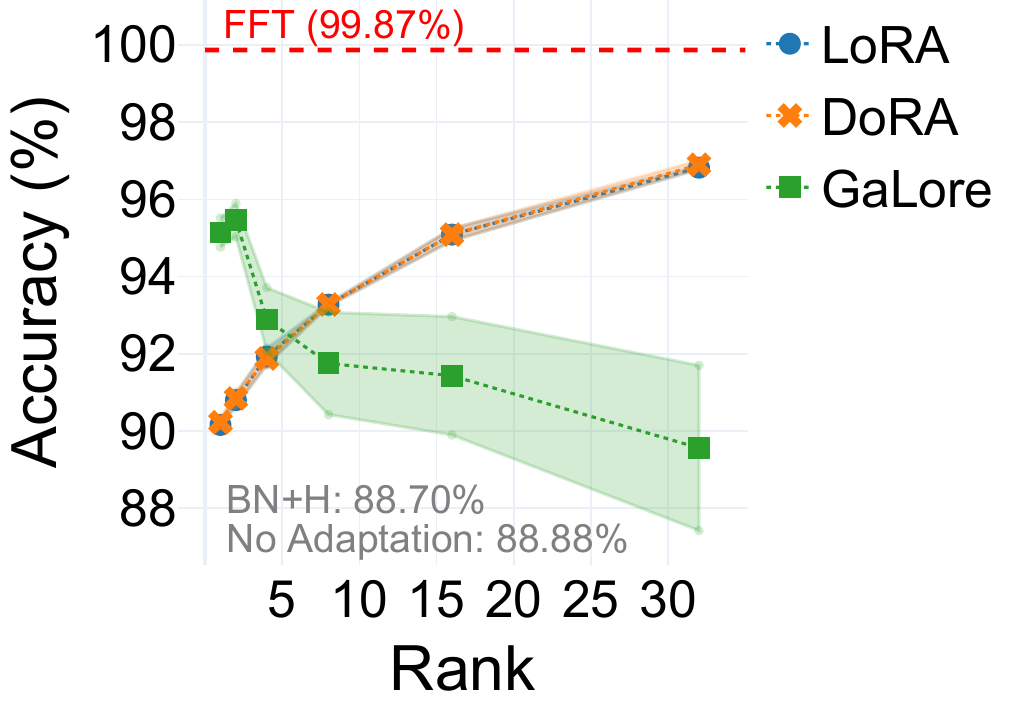}
    \label{fig:rank-acc-mnv2-1-f}
}
\caption{\textbf{Impact of different ranks on adaptation accuracies.} Two pre-trained (\ie ImageNet and CIFAR10) MobileNetV2 models are fine-tuned for five epochs on three different datasets (\ie CIFAR10-C Brightness (br), CIFAR10-C Impulse noise (in), and \visualwakewords) by varying the \peft methods ranks. \peft methods' performance is influenced by the initial task accuracy of each pre-trained model.}
\label{fig:rank-acc-mnv2-1}
\end{figure*}

\textbf{Impact of the rank.} Theoretically, a \peft method’s rank determines its learning capacity in the low-rank space, with higher ranks better approximating FFT performance~\citep{hu2021lora, zhao2024galorememoryefficientllmtraining}. However, as shown in \figref{fig:rank-acc-mnv2-1}, fine-tuning a pre-trained model for 5 epochs reveals that higher rank does not always improve accuracy.

Unlike LLM results in~\citet{zhao2024galorememoryefficientllmtraining}, we find that for tasks where the pre-fine-tuning accuracy is high (\figref{fig:rank-acc-mnv2-1-a} and \figref{fig:rank-acc-mnv2-1-f}), the accuracy of \galore may degrade with a higher rank setting. In these cases, \lora and \dora achieve up to 6\% higher accuracy compared to \galore. We conjecture that for these fine-tuning tasks, a lower rank setting provides a smoother gradient landscape that is easier to optimize, leading to better model performance.

When the pre-trained model performs poorly on a fine-tuning task, adapter-based methods like \lora and \dora often struggle to adapt (\figref{fig:rank-acc-mnv2-1-e}). In such cases, \galore outperforms them by up to 50\% at low ranks, benefiting from its full-rank weight updates. We hypothesize that the combined gradient rank of \lora and \dora is insufficient for small $r$, particularly when the fine-tuning task diverges significantly from the pre-training objective. This idea is supported by a similar effect observed in federated learning scenarios in~\citet{babakniya2023slorafederatedparameterefficient}, where increasingly diverse data distributions across clients also increased the accuracy gap between \lora and \fullfinetuning. 

Results with CIFAR-10 pretraining follow the same trends. Although \citet{liu2024dora} reports \dora outperforming \lora by up to 37\% for $r < 16$, we observe no such gain, as both perform similarly across tasks. Further analysis is left for future work.

\textbf{Summary.} 
Overall, the \peft methods \lora, \dora, and \galore significantly outperform \bnh in terms of accuracy, especially on challenging fine-tuning tasks. \galore demonstrates robust accuracy and requires, on average, approximately 2$\times$ fewer iterations during fine-tuning compared to \lora, \dora and \bnh.
Although \lora offers a better trade-off between resources and accuracy, with particularly low \flops consumption on models employing \dscs and significantly improved memory efficiency on standard convolution \cnns, it comes at the cost of longer training times and less robust accuracy scores on hard fine-tuning tasks.
While \dora introduces a memory overhead compared to \fullfinetuning, it does not show improved performance over \lora in any of our experiments, making it less efficient than \lora for \cnns optimized for edge devices.

\section{Related Work}
\textbf{Test-time Adaptation on the Edge.} 
Test-time adaptation (\tta) has emerged as a technique for dynamically adjusting model's parameters to the incoming test data stream to address domain shifts~\citep{niu2024test, wang2020tent}. \tta enables deep models to adjust their predictions based on the characteristics of the incoming test data, which may differ from the training data~\citep{iwasawa2021test}. While \tta is extensively employed to adapt deep models at the edge, it presents several limitations, including representation collapse due to overfitting on the test data~\citep{press2024entropy} and the inability to handle open set domain adaptations~\citep{panareda2017open}. Furthermore, \tta model's weight update exhibits memory usage similar to full fine-tuning, \ie infeasible at the edge, since the model needs to backpropagate through its whole architecture to compute the gradient~\citep{jiatinytta} and store it in the optimizer memory buffer. \peft methods mitigate these limitations by learning a set of low-rank external matrices~\citep{hu2021lora}, or by fine-tuning the model with a memory-efficient low-rank projection of the gradient~\citep{zhao2024galorememoryefficientllmtraining} and therefore reducing the required optimizer memory.     

\textbf{Parameter-Efficient Fine-Tuning for \llms.}
\peft methods have drawn attention for fine-tuning large language models (\llms) without incurring the prohibitive computational or memory costs of standard fine-tuning methods. \peft methods introduce a small number of additional parameters to be fine-tuned, \ie through low-rank decomposition and specialized adapters, while keeping the model weights frozen or updating the weights through a low-rank projection of the gradients. While these methods have been studied for \llms fine-tuning~\citep{han2024parameter}, analysis of their performance for on-device deep model updates is lacking. In contrast to previous works, we explore the performance of \peft methods for optimized deep learning models for edge devices, and compare their performance and computational cost for different downstream tasks. 

\section{Conclusion, Limitations, and Outlook}
This study benchmarks parameter-efficient fine-tuning (\peft) methods on CNN architectures for edge devices, revealing distinct trade-offs between accuracy and resource efficiency. While \lora achieves the best balance between performance and computational cost in most scenarios, \dora's additional memory overhead limits its applicability in resource-constrained settings. \galore demonstrates robust accuracy but incurs higher computational complexity due to \singularvaluedecomp-based updates. Across architectures using depthwise-separable convolutions, \peft methods are only half as memory-efficient as reported for \llms, with adapter-based methods achieving up to 95\% \flops reduction compared to full fine-tuning. These findings provide actionable insights into selecting \peft methods for edge deployments, depending on hardware constraints and application needs.

\textbf{Limitations.}
This study does not include on-device profiling, limiting realism for specific deployment scenarios. Comparisons are based on fixed hyperparameters without extensive tuning, and quantization is not considered. While focused on \cnns, generalization to other edge-relevant architectures like lightweight transformers remains unexplored. These aspects are left for future work.

\textbf{Acknowledgments.}
This research was funded in part by the Austrian Science Fund (FWF) within the DENISE doctoral school (grant number DFH 5). The results presented in this paper were computed using the computational resources of the HLR resources of the Zentralen Informatikdienstes of Graz University of Technology.

\bibliography{arxiv}
\clearpage
\section*{Appendix}
\input{arxiv_appendix}

\end{document}

%% file: arxiv_appendix.tex
\begin{table}[!htbp]
    \centering
    \renewcommand{\arraystretch}{1.2} 
    \setlength{\tabcolsep}{8pt} 
    \caption{\textbf{Hyperparameters.} Baseline hyperparameters for the analyzed \peft methods, consistent with~\cite{hu2021lora, liu2024dora, zhao2024galorememoryefficientllmtraining}.}
    \label{tab:hyperparams}
    \begin{tabular}{l|c|p{0.55\linewidth}} 
        \hline
        \textbf{Parameter} & \textbf{Value} & \textbf{Description} \\ 
        \hline
        $r_{\text{\lora}}$ & 4 & Rank of the low-rank adapter. \\ 
        $\alpha_{\text{\lora}}$ & 4 & Influence of the adapter result scaled by $\alpha / r$. \\ 
        \hline
        $r_{\text{\dora}}$ & 4 & Rank of the low-rank adapter. \\ 
        $\alpha_{\text{\dora}}$ & 4 & Influence of the adapter result scaled by $\alpha / r$. \\ 
        \hline
        $r_{\text{\galore}}$ & 4 & Gradient approximation rank. \\ 
        $\text{scale}_{\text{\galore}}$ & 0.25 & Gradient approximation scaling factor. \\ 
        $T$ & 200 & Subspace update frequency. \\ 
        \texttt{proj\_type} & \texttt{std} & Projection of the low-rank gradient. \\ 
        \hline
    \end{tabular}
\end{table}

\begin{figure*}[!htbp]
    \centering
    \subfigure[MobileNetV2, FLOPs]{
        \label{fig:rank-flops-mnv2}
        \includegraphics[width=0.21\linewidth, trim={0cm 0cm 4.3cm 0cm},clip]{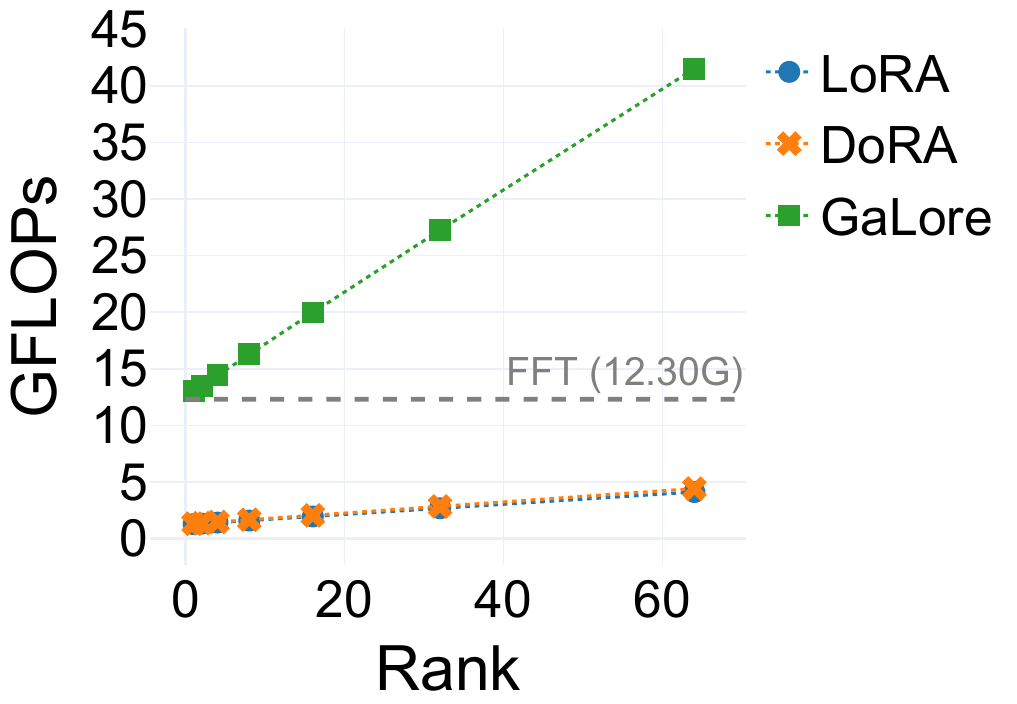}
    }
    \subfigure[ResNet-18, FLOPs]{
        \label{fig:rank-flops-plnm}
        \includegraphics[width=0.21\linewidth, trim={0cm 0cm 4.3cm 0cm},clip]{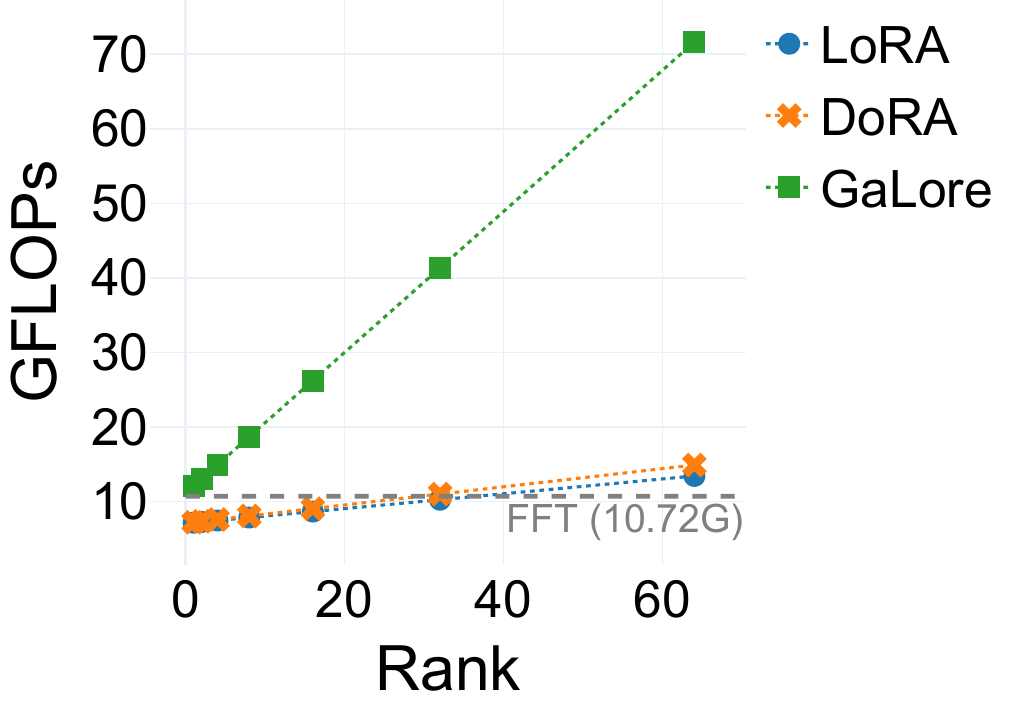}
    } 
    \subfigure[MobileNetV2,Memory]{
        \label{fig:rank-flops-mnv3}
        \includegraphics[width=0.21\linewidth, trim={0cm 0cm 5cm 0cm},clip]{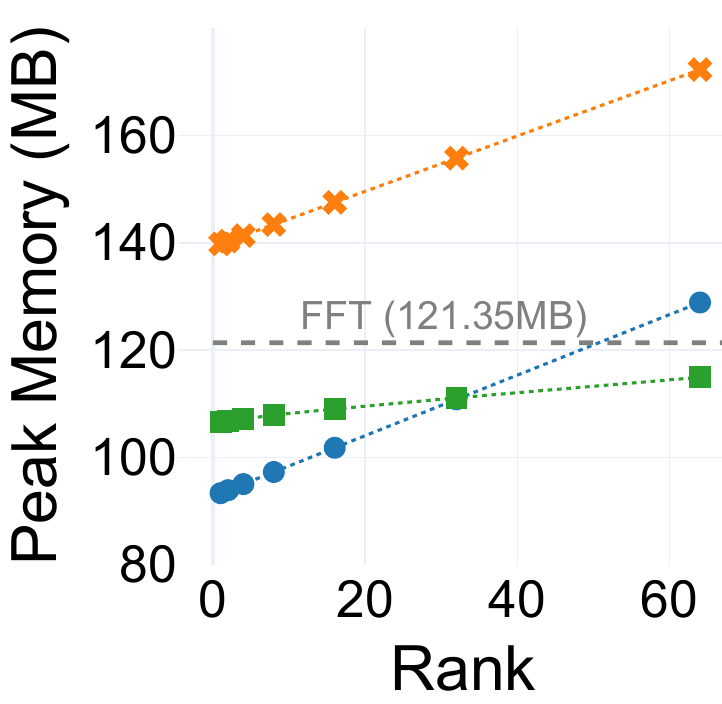}
    }
    \subfigure[ResNet-18, Memory]{
        \label{fig:rank-flops-rn18}
        \includegraphics[width=0.26\linewidth, trim={0cm 0cm 0cm 0cm},clip]{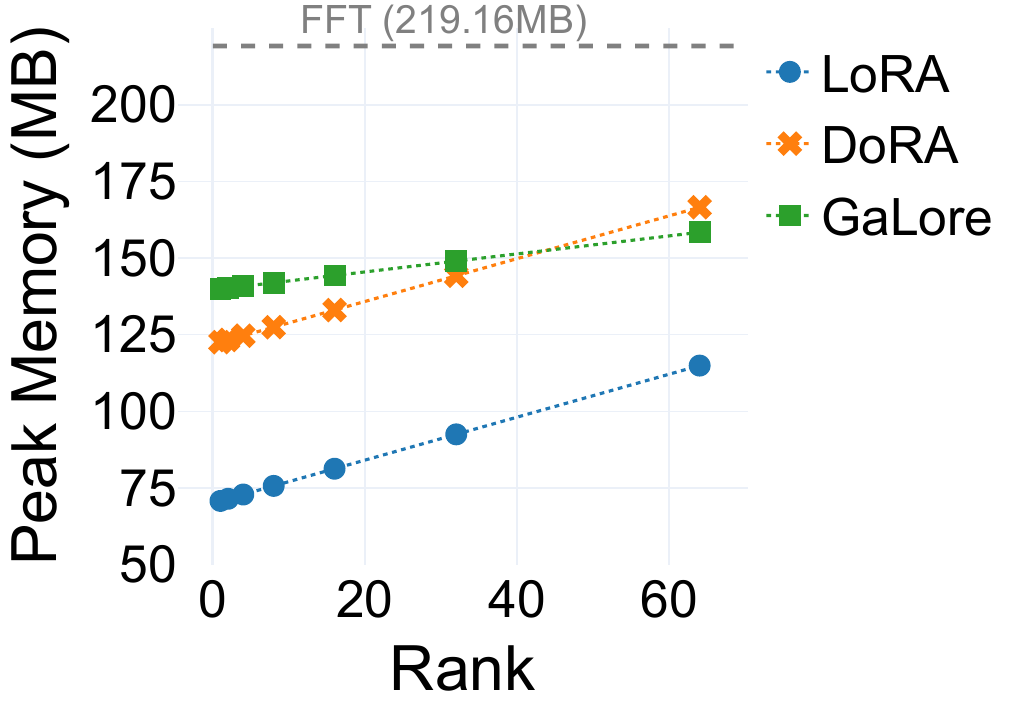}
    }
    \caption{\textbf{Impact of rank on \peft resource efficiency.} The memory consumption required for a single $224\times224$ image depends on the rank. \flops and memory consumption scale linearly with increasing rank. The \flops consumption for \galore increases approximately 9 times faster than for \lora and \dora, while memory consumption shows the opposite trend, with \lora and \dora increasing 2-4 times more steeply than \galore, depending on the model architecture.}
    \label{fig:rank-resources}
\end{figure*} 